\documentclass[twocolumn, switch]{article} 

\usepackage{preprint}
\usepackage{enumitem}
\usepackage{caption}
\usepackage{subcaption}
\usepackage{graphicx}

\setlist[itemize]{leftmargin=*}
\setlist[enumerate]{leftmargin=*}

\usepackage{amsmath, amsthm, amssymb, amsfonts}

\usepackage{natbib}
\usepackage[utf8]{inputenc}	
\usepackage[T1]{fontenc}	
\usepackage{xcolor}		
\usepackage[colorlinks = true,
            linkcolor = purple,
            urlcolor  = blue,
            citecolor = cyan,
            anchorcolor = black]{hyperref}	
\usepackage{booktabs} 		
\usepackage{nicefrac}		
\usepackage{microtype}		
\usepackage{lineno}		
\usepackage{float}			

\usepackage{lipsum}		

\usepackage{newfloat}
\DeclareFloatingEnvironment[name={Supplementary Figure}]{suppfigure}
\usepackage{sidecap}
\sidecaptionvpos{figure}{c}
\usepackage{comment}

\usepackage{titlesec}
\titlespacing\section{0pt}{12pt plus 3pt minus 3pt}{1pt plus 1pt minus 1pt}
\titlespacing\subsection{0pt}{10pt plus 3pt minus 3pt}{1pt plus 1pt minus 1pt}
\titlespacing\subsubsection{0pt}{8pt plus 3pt minus 3pt}{1pt plus 1pt minus 1pt}

\date{}
\title{NGD converges to less degenerate solutions than SGD}

\usepackage{authblk}
\usepackage{textcomp}

\author{Moosa Saghir\textsuperscript{*}}
\author{N. R. Raghavendra\textsuperscript{*}}
\author{Zihe Liu\textsuperscript{*}}
\author{Evan Ryan Gunter}

\affil{Cambridge AI Safety Hub}

\begin{document}

\twocolumn[ 
  \begin{@twocolumnfalse} 

    \maketitle
    
    \begin{abstract}

The number of free parameters, or \textit{dimension}, of a model is a straightforward way to measure its complexity: a model with more parameters can encode more information. However, this is not an accurate measure of complexity: models capable of memorizing their training data often generalize well despite their high dimension \cite{doubledescent}. \textit{Effective dimension} aims to more directly capture the complexity of a model by counting only the number of parameters required to represent the functionality of the model \citep{effective_dimension_evan}. Singular learning theory (SLT) proposes the learning coefficient \( \lambda \) as a more accurate measure of effective dimension \citep{Watanabe2009}.
By describing the rate of increase of the volume of the region of parameter space around a local minimum with respect to loss, $\lambda$ incorporates information from higher-order terms. We compare \( \lambda \) of models trained using natural gradient descent (NGD) and stochastic gradient descent (SGD), and find that those trained with NGD consistently have a higher effective dimension for both of our methods: the Hessian trace \( \text{Tr}(\mathbf{H}) \), and the estimate of the local learning coefficient (LLC) $\hat{\lambda}(w^*)$.
        \end{abstract}
    \vspace{0.35cm}

  \end{@twocolumnfalse} 
] 



\renewcommand{\thefootnote}{\fnsymbol{footnote}}
\footnotetext[1]{Equal contribution}
\renewcommand{\thefootnote}{\arabic{footnote}}


\section{Introduction}
\subsection{SLT Motivation}

Model complexity measures the amount of information captured by models, crucial to understanding the learning and generalization behaviour of models. The nominal parameter count has traditionally been a proxy for complexity, however does not accurately reflect the amount of information captured. It often overestimates complexity, as demonstrated by pruned models which achieve similar performance to the original models while containing a smaller amount of information \citep{blalock2020stateneuralnetworkpruning}.

The effective dimension more accurately captures model complexity. Given a model with $N$ parameters and $N_\text{free}$ degrees of freedom in the parameterisation, we could theoretically remove these redundant degrees of freedom to obtain a functionally identical model with only $N - N_\text{free}$ parameters; so, the effective dimension of the model is $N - N_\text{free}$ \citep{bushnaq2024usingdegeneracylosslandscape}. 
Multiple candidates such as \( \text{rank}(\mathbf{H}_{L}) \) of the Hessian of loss or the 'effective parameter count' raised in \citep{BayesianModelMackay} estimate the effective dimension. \citep{hessian_zero_eigenvalues} shows that most of the eigenvalues of \( \mathbf{H}_{L} \) are densely concentrated near zero \footnote{In practice, $\mathbf{H}$  will almost certainly have full rank. Eigenvalues of $\mathbf{H}$ may be very small, but not exactly 0. } for models trained using SGD.  In other words, there are a large number of directions (equivalently parameters) in which the loss changes little as we traverse the loss landscape in that direction. Only the number of large eigenvalues contribute to the effective dimension; we should not include these small eigenvalues in the dimension count. A model is \textit{degenerate} due to these near-zero eigenvalues, and a more degenerate model has more of these near-zero eigenvalues.

Note that the Hessian rank only provides us with a second-order approximation of the geometry of the loss landscape:

\begin{equation}
\begin{aligned}
L(\mathbf{w}) = L(\mathbf{w}^*) + \nabla L(\mathbf{w}^*)^\top (\mathbf{w} - \mathbf{w}^*)+{} \\
\frac{1}{2}(\mathbf{w} - \mathbf{w}^*)^\top \mathbf{H}(\mathbf{w}^*)(\mathbf{w} - \mathbf{w}^*) + O(\|\mathbf{w} - \mathbf{w}^*\|^3)
\end{aligned}
\end{equation}

where $w^*$ is a local minimum. Higher order terms in \( L \) are not considered in the Hessian. \( \mathbf{H} = \mathbf{0} \) does not necessarily imply flatness in all directions, because the loss might have non-zero terms higher than the second order.

SLT proposes the LLC $\lambda(w^*)$
to be the correct measure of a model's effective dimension \footnote{SLT proposes that the global learning coefficient $\lambda$ fully captures the effective dimension, but this is intractable to calculate. We focus on estimating the local learning coefficient $\lambda(w^*)$}.
Like the Hessian rank, ${\lambda}(w^*)$ tells us about the degeneracy of the solution; a lower ${\lambda}(w^*)$ implies a broader, more degenerate solution. Crucially, ${\lambda}(w^*)$ takes into account all higher order terms in the Taylor expansion of \( L \).

Previous research has used the parameter count for model selection. Specifically, \citep{BIC} uses the Bayesian Information Criterion (BIC):
\begin{equation}
\text{BIC} = n L_n(w^*) + \frac{d}{2} \log{(n)}
\end{equation}
where $ n $ is the number of training samples, \( L_n(w^*) \) is the empirical loss for a minimum $w^*$ maximizing the likelihood, and \( d \) is the parameter count. This expression displays an \textit{accuracy-complexity trade-off} because it consists of a loss term \(n L_n(w^*)\) and a term \( \frac{d}{2} log{(n)} \) penalising complexity. A model with slightly worse empirical loss but significantly fewer parameters may generalize better than an ``overfitted'' model which achieves better performance on the training set but uses more parameters.

Under SLT, the BIC is generalised to the Widely Applicable Bayesian Information Criterion (WBIC) where:

\begin{equation}
\text{WBIC} = n L_n(w^*) + \lambda \log{(n)}
\end{equation}

and $\lambda < d/2$ in general. SLT proposes the same accuracy-complexity trade-off, but with $\lambda$ measuring complexity instead of $d$.

\subsection{Natural Gradient Descent}

The LLC $\lambda(w^*)$ has been used to analyse the complexity of models \citep{lau2023quantifying}. However, it remains an open question empirically whether SLT can be applied to analysing the complexity of NGD and SGD. In particular, \citep{PhysRevLett.130.237101} suggests that SGD biases models towards \textbf{more degenerate singularities} in the loss landscape. A degenerate point is where the Fisher Information Matrix, denoted by \( \mathbf{F} \), is singular \footnote{In this paper, we use singularities and degenerate points interchangeably. More singular or degenerate points can either mean the small eigenvalues are closer to 0, or there are more small eigenvalues} (and so, \(\mathbf{H}\) is also singular because \( \mathbf{F} = \mathbf{H}\) at a local minimum) \citep{lau2023quantifying}. In practice, we see that singular points have at least one small eigenvalue (close to 0), and more degenerate points have more small eigenvalues hence a lower approximate $\text{rank}(\mathbf{H})$. We might expect these points to have a lower effective dimension $\lambda$ as they have more $N_\text{free}$ parameters, and have a smaller curvature in the loss landscape.

In this paper, we analyse the complexity of models trained using NGD and SGD.
In an SGD update step, a constant multiple of the current negative gradient is added to the current parameters:
\begin{equation}
w_{n+1} = w_{n} - \eta  \nabla L
\end{equation}

This works well for loss landscapes which curve slowly relative to the size of the update steps, and which do not have flat areas away from local optima (saddle points).
The first of these conditions implies that when the gradient is large, the current location in the loss landscape is far from a local optimum where the gradient is 0, so we should take a large step down the slope.
The second condition implies that when the gradient is small, the current location in the loss landscape is close to a local optimum, so we should take only small steps to avoid overshooting the minimum, and once the steps have become very small we must have a model that performs almost identically to the model where the gradient is exactly zero.

However, these conditions do not hold in real loss landscapes, which may be sharply curved and tend to have far more saddle points than minima \citep{saddlepoints}.
NGD attempts to address these problems by scaling the gradient based on the curvature in each direction \citep{ngd_original_paper}:

\begin{equation}
w_{n+1} = w_{n} - \eta \mathbf{F^{-1}} \nabla L
\end{equation}

This is equivalent to stretching the loss landscape such that the magnitude of the curvature is the same in every direction, taking an SGD step in this stretched landscape, and translating this back to the equivalent point on the original landscape (i.e. the pullback of the gradient by \( \mathbf{F^{-1}} \)). Hence, NGD updates slowly in high-curvature regions and quickly in smoother regions, which can help avoid oscillatory behaviour and achieve faster convergence.

We use the NGD implementation in \citep{yao2020pyhessian}. Their implementation smooths the Fisher matrix with the term $\kappa \mathbf{I}$, such that the smoothed Fisher matrix $\mathbf{F_s}$ is:

\begin{equation} \label{F_s}
\mathbf{F_s} = \mathbf{F} + \kappa \mathbf{I}
\end{equation}

where, for a model with $d$ parameters, $\kappa$ is given by:

\begin{equation} \label{alp_eps}
\kappa = \frac{\alpha}{d} \ \text{max}(\text{trace}(\mathbf{F}), \epsilon)
\end{equation}
The smoothing is necessary to ensure that NGD converges, and to avoid computational problems with extremely small eigenvalues. Note that in practice we use \( \mathbf{F_s^{-1}}\) instead of \( \mathbf{F^{-1}} \), so we can still take update steps at singularities. As \( \kappa \to \infty \), NGD effectively becomes SGD because \( \mathbf{F_s} \) becomes the scaled identity matrix.

We hypothesise that NGD, with moderate smoothing, will converge to solutions with a higher \( \lambda \) than SGD, because of the \( \mathbf{F^{-1}} \) term causing NGD to avoid flatter, highly degenerate regions. In these regions, many of the eigenvalues of $\mathbf{F}$ become extremely small, hence the point is highly degenerate. The determinant of $\mathbf{F}$ becomes very low, possibly causing a large update step to be taken in these degenerate regions.

\subsection{Theoretical framework of SLT}
\subsubsection{Background}
SLT is conducted within the Bayesian paradigm, where we have some data $D_n$ with \( n \) samples, such that:

\begin{equation}
D_n = \{(X_1, Y_1), (X_2, Y_2), \dots, (X_n, Y_n)\}
\end{equation}

where $D_n$ is independent and identically distributed according to $q(y,x)=q(y|x)q(x)$. Let $W$ be the space of all possible parameters $w$. We consider the following distributions:

\begin{itemize}
\item $p(w)$: Prior probability density function, $w \in W$
\item $q(x)$: True distribution of inputs
\item $q(y|x)$: True conditional distribution of outputs
\item $p(y|x,w)$: Model to estimate $q(y|x)$
\end{itemize}

We aim to find $w$, parameterizing $p(y|x,w)$, that accurately describes the relationship between $X$ and $Y$ in $q(y|x)$.

The Kullback-Leibler (KL) divergence $D_{KL}$ between the true distribution $q(y,x)$ and model $p(y|x,w)$ is

\begin{equation}
D_{KL} \left[ q(y,x) \parallel p(y|x,w) \right] = \int\!\!\!\int q(y,x) \log \frac{q(y|x)}{p(y|x,w)} \,dx\,dy
\end{equation}

and this only depends on our parameters $w$:

\begin{equation}
K(w) = D_{KL} \left[ q(y,x) \parallel p(y|x,w) \right]
\end{equation}

We define the empirical loss \( L_n(w) \) as a measure of error, which is just the average negative log likelihood:
\begin{equation} 
\begin{aligned}
    L_n(w) =  -\frac{1}{n}\sum_{i=1}^{n} \log p(y_i \vert x_i,w) \\
\end{aligned}
\end{equation}

For the loss $L(w)$ in the asymptotic case $n \to \infty$,

\begin{equation}
K(w) = L(w) - S
\end{equation}

where S is a constant and $K(w) \geq 0 \text{ for all } w \in W$. We can think of $K(w)$ as a "loss metric" between the truth and the model since minimizing $K(w)$ is akin to minimizing $L(w)$, so that our model describes the truth:

\begin{equation}
W_0 = \{w \in W \mid K(w) = 0\} = \{w \in W \mid p(y|x,w) = q(y|x)\}
\end{equation}

\subsubsection{The LLC $\lambda(w^*$)}

We formally introduce the LLC $\lambda(w^*)$. At a particular local minimum $w^*$ and loss $K(w^*)$, $V_{w^*}(\epsilon)$ is the volume of parameter space with loss $K(w) \leq K(w^*) + \epsilon$  in a surrounding neighbourhood $W_{w^*} \subset W$. \footnote{We obtain the global learning coefficient $\lambda$ using the global minimum $w_0$ instead of a local minimum $w^*$, where $w_0 \in W_0$ and $K(w_0)=0$}

\begin{equation}
V_{w^*}(\epsilon) = \int_{w \, \in \, W_{w^*} \, : \, K(w) \, \leq \, K(w^*) + \epsilon} \! dw
\end{equation}

The LLC is then

\begin{equation}
\lambda(w^*) = \lim_{\epsilon \to 0} \left[ \epsilon \frac{d}{d\epsilon} \log V_{w^*}(\epsilon) \right]
\end{equation}

and can be thought of as how fast this volume $V$ in a basin scales when the error threshold $\epsilon$ is raised. Under some assumptions, we see that

\begin{equation}
V(\epsilon) \propto \epsilon^{\lambda} \quad \text{as} \quad \epsilon \to 0.
\end{equation}

Thus, for arbitrarily small $\epsilon$, a higher $\lambda$ implies $V$ grows slowly inside the basin, whilst a lower $\lambda$ implies $V$ grows quickly.

\subsubsection{Generalization}

In the asymptotic case as $n \to \infty$, \citep{Watanabe2009} shows that the Bayesian generalization error is

\begin{equation}
    G_n = L(w_0) + \frac{\lambda}{n} + o\left(\frac{1}{n}\right) = \mathbb{E}_{D_n}[K(w)]
    \label{eq:G_n}
\end{equation}

We see that the generalization loss displays an accuracy-complexity trade-off similar to the one discussed in the BIC. \footnote{While the global minimum $w_0$ is used here, we extend this idea to local minimums $w^*$} Less complex minima have a lower $\lambda$, and we expect better generalization performance. This is the more degenerate, singular case that has a lower effective dimension. For 2 different $w^*$ giving the same accuracy, the one with a lower $\lambda(w^*)$ will be expected to generalize better.

\section{Methodology}

\subsection{Summary}

Fully connected feed-forward neural networks (FFNN) and Convolutional Neural Nets (CNN) following the Lenet-5 architecure \citep{lenet} are independently trained with varying numbers of hidden nodes per layer and hidden layers on the MNIST and Fashion MNIST benchmark datasets. We approximate the trace of the Hessian of the model at each epoch during training using the \texttt{PyHessian} library \citep{yao2020pyhessian}. $\lambda(w^*)$ and the WBIC are estimated at each epoch using the \texttt{devinterp} library \citep{devinterp2024} using the stochastic gradient LLC estimator.

The following experiments were conducted:
\begin{enumerate}
    \item Train models independently using SGD and NGD, and compare the evolution of $\hat{\lambda}$, the WBIC, and $\text{Tr}(\mathbf{H})$
    \item Train models with NGD and SGD, with different values of smoothing constants $\alpha$ and $\epsilon$ for NGD
    \item First train one model with SGD, then when loss and LLC have stabilized, train two models from the same state using SGD and NGD
\end{enumerate}

Throughout training, the following quantities are computed:
\begin{itemize}
    \item \textbf{Training and validation loss}: Negative Log Likelihood loss on train and validation datasets respectively
    \item \textbf{Update norm}: mean over batches of the norm of the update vector calculated in each epoch
\end{itemize}

\subsection{Estimating the local learning coefficient $\hat{\lambda}(w^*)$ and the WBIC}

We estimate the LLC by sampling around the local neighbourhood of our point of interest. In \citep{lau2023quantifying}, the free energy \( F_n \) is estimated using the WBIC, which is estimated as:

\begin{equation}
WBIC = \hat{F_n} := \mathbb{E}_{w|w*}^{\beta}[nL_n(w)]
\end{equation}
Rearrangement of the analytic form of the WBIC naturally leads to the \textit{LLC estimator}:

\begin{equation}
\hat{\lambda}(w^*) = \beta \left( \mathbb{E}_{w|w^*}^{\beta} \left[ nL_n(w) \right] - nL_n(w^*) \right)
\end{equation}

\begin{itemize}
    \item $\hat{\lambda}(w^*)$: The estimate of the LLC at $ w^* $.
    \item $\beta$: The inverse temperature; analogous to thermodynamics, higher $\beta$ reduces the entropy in the posterior distribution $p(w|D_n)$, making it more sharply peaked. This is typically set to \( \log{(n)}\).
    \item $n$: The size of the training dataset.
    \item $\mathbb{E}_{w|w^*}^{\beta} \left[ nL_n(w) \right]$: The expected value of $n L_n(w)$ for $w$ in the local neighbourhood near $w^*$, under the \textit{localized tempered posterior distribution}, such that:
    \begin{equation}
    p(w|D_n, w^*) \propto e^{-\beta nL_n(w) - \gamma ||w - w^*||^2}
    \end{equation}
    
    where $\gamma$ is the localization; it flattens the posterior distribution for $w$ far away from $w^*$, ensuring that the estimate of $\lambda(w^*)$ is local.
\end{itemize}

\citep{lau2023quantifying} also uses SGLD in the update equation for sampling:

\begin{equation}
\Delta w_t = \frac{\epsilon}{2} \left( \beta n \sum_{i=1}^{m} \nabla \log p(y_i | x_i, w_t) + \gamma(w^* - w_t) \right) + \mathcal{N}(0, \varepsilon)
\end{equation}

\begin{itemize}
    \item $\epsilon$ : the step size.
    \item $m$ : the batch size.
    \item $\gamma$ : the localization / confinement strength. A large $\gamma$ means that for $w$ that is far from $w^*$, the update step will be large in the direction of $w^*$.
\end{itemize}

We calculate $\hat{\lambda}$ for the model at each epoch and track its evolution. While theoretically $\lambda(w^*)$ is only valid for models that have converged to minimums, we still find experimentally meaningful results measuring $\hat{\lambda}(w)$ for $w$ throughout training.

\section{Experiments and results}

\subsection{NGD solutions have higher $\hat{\lambda}$ and $\text{Tr}(\mathbf{H})$ than SGD}
\begin{figure}[htbp]
    \centering
    \begin{subfigure}[t]{0.45\textwidth}
        \centering
        \includegraphics[width=0.83\textwidth]{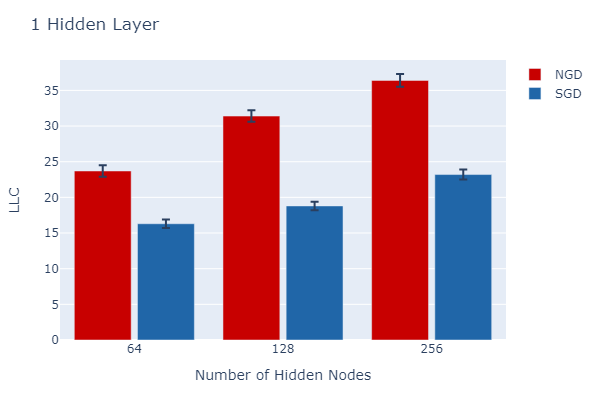}
        \caption{Mean $\hat{\lambda}$ and error bars for models with 1 hidden layer.}
        \label{fig:1HL}
    \end{subfigure}
    \hfill
    \begin{subfigure}[t]{0.45\textwidth}
    \centering
        \includegraphics[width=0.83\textwidth]{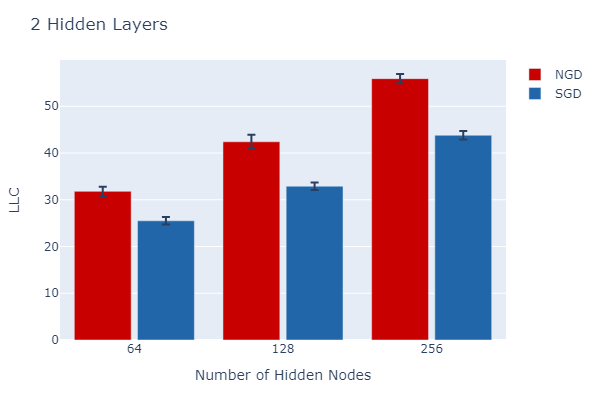}
        \caption{Mean $\hat{\lambda}$ and error bars for models with 2 hidden layers.}
        \label{fig:2HL}
    \end{subfigure}
    
    \caption{NGD solutions have higher $\hat{\lambda}$, with highest t-value of \(1.9^{-31}\), over a range of model sizes. Hyperparameters used: \( \alpha = 10^{-2}\), \( \text{learning rate} = 10^{-2}\), \( \epsilon = 10^{-10}\), \( \text{batch size} = 128\).}
    \label{fig:combined_models}
\end{figure}

\begin{figure}[t]
    \centering
    \begin{subfigure}[t]{0.45\textwidth}
    \centering
        \includegraphics[width=0.84\textwidth]{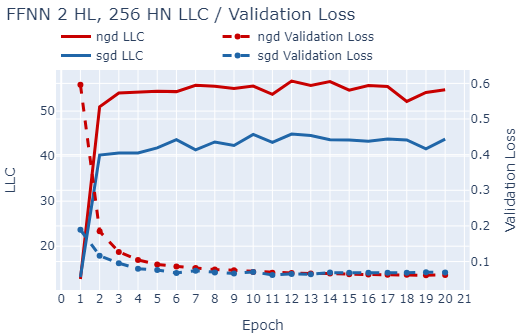}
        \caption{$\hat{\lambda}$ is higher for NGD across epochs.}
        \label{fig:llc}
    \end{subfigure}

    \vspace{0.5cm}
    
    \begin{subfigure}[t]{0.45\textwidth}
    \centering
        \includegraphics[width=0.84\textwidth]{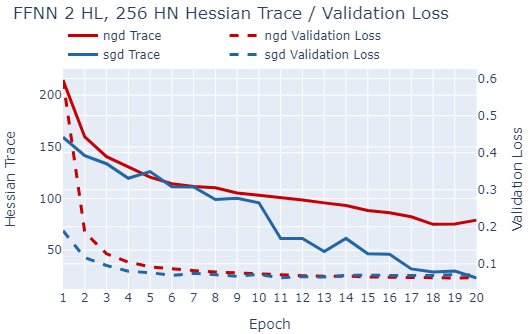}
        \caption{\( \text{Tr}(\mathbf{H})\) is higher for NGD over epochs.  }
        \label{fig:hess}
    \end{subfigure}
    
    \caption{Hyperparameters used: \( \alpha = 10^{-1}\), \( \text{lr} = 10^{-2}\), \( \epsilon = 10^{-10}\), \( \text{batch size} = 128\).}
    \label{fig:combined_llc_hess}
\end{figure}

Our results show that the LLC $\hat{\lambda}$ is higher in NGD-trained models across a range of model sizes (Figure \ref{fig:combined_models}). Across epochs, the LLC for NGD and SGD plateau, while the Hessian trace continues to decrease (Figure \ref{fig:combined_llc_hess}).

\subsection{Reducing Fisher matrix smoothing increases the LLC of NGD}

In Equation \ref{F_s}, we see that \( \kappa \) controls the amount of smoothing applied to \( \mathbf{F} \). In the limit as \( \kappa \to \infty \), \( \mathbf{F_s} \to \kappa \mathbf{I} \), so NGD effectively becomes SGD with learning rate \( \frac{\eta}{\kappa} \).

\begin{figure}[h]
    \centering
    \begin{subfigure}[t]{0.45\textwidth}
        \includegraphics[width=0.9\textwidth]{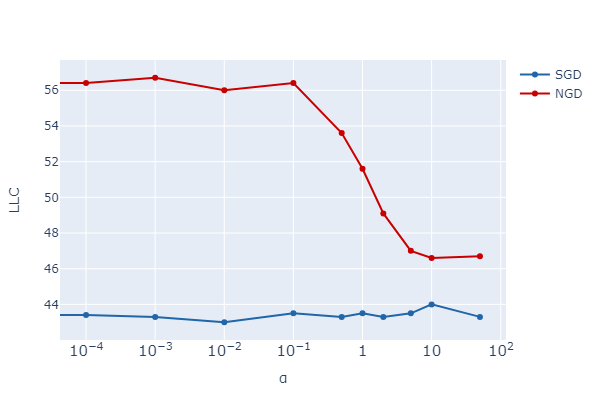}
        \caption{Reducing \( \alpha \) increases $\hat{\lambda}_{NGD}$.}
        \label{fig:alpha}
    \end{subfigure}

    \begin{subfigure}[t]{0.45\textwidth}
        \includegraphics[width=0.9\textwidth]{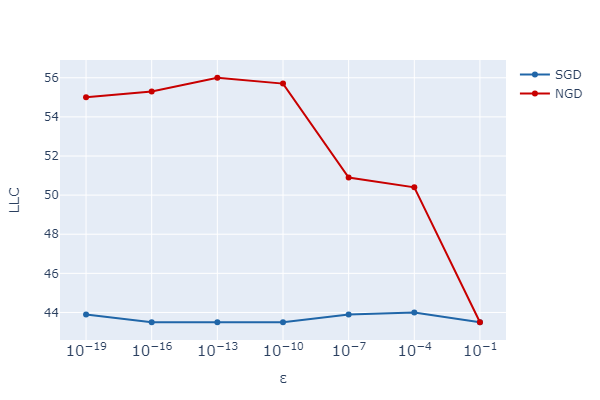}
        \caption{{Reducing \( \epsilon \) increases $\hat{\lambda}_{NGD}$.}}
        \label{fig:epsilon}
    \end{subfigure}
    
    \caption{}
    \label{fig:combined_alpha_epsilon}
\end{figure}

From Equation \ref{alp_eps}, \( \kappa \) is directly proportional to \( \alpha \), and \( \frac{\alpha \epsilon}{d}\) is a lower bound on \( \kappa \). Increasing either \( \alpha \) or \( \epsilon \) will increase the smoothing factor \( \kappa \) and bring NGD closer to SGD. Therefore, we would expect \( \lambda_{NGD} \) to approach \( \lambda_{SGD} \) for large \( \alpha \) and \(\epsilon\), because the update steps of NGD become less volatile near a degenerate singularity. Experimentally, we observe that increasing \( \alpha \) and \( \epsilon \) does indeed reduce the LLC of NGD in Figure \ref{fig:alpha} and Figure \ref{fig:epsilon}.

\subsection{NGD escaping a basin with highly degenerate singularities}

Initially, we train either a FFNN (2 layers, 256 hidden nodes each) on the Fashion-MNIST and MNIST datasets using SGD. After the initial sharp drop in loss, and when the LLC has stabilized, we then train 2 models from the same state, one using NGD and the other continuing with SGD. We measure the average update step size in each epoch, where we use $|| \nabla L ||$ for SGD and $|| \mathbf{{F_s}^{-1}} \nabla L ||$ for NGD. After the split, we also try different learning rates of NGD to see its effect on $\hat{\lambda}_{NGD}$.

\begin{figure}[htbp]
    \centering
    \begin{subfigure}[t]{0.45\textwidth}
    \centering
        \includegraphics[width=0.9\textwidth]{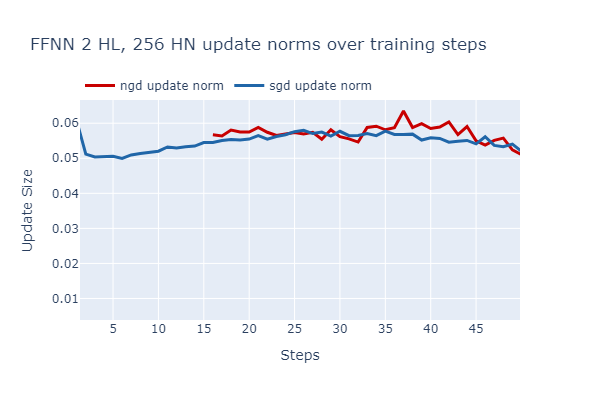}
        \caption{Update step size changes little after splitting}
        \label{fig:fashion_update}
    \end{subfigure}

    \begin{subfigure}[t]{0.45\textwidth}
    \centering
        \includegraphics[width=0.9\textwidth]{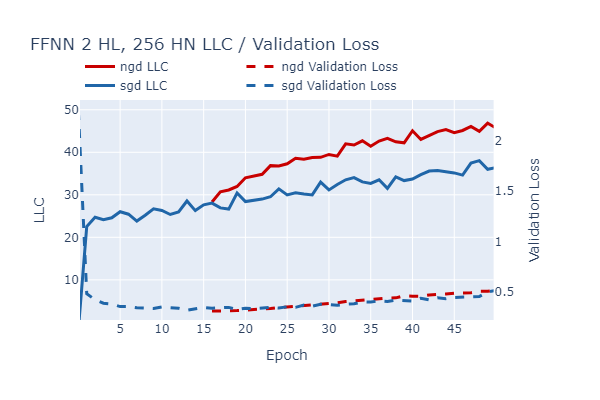}
        \caption{$\hat{\lambda}_{NGD}$ steadily increases after splitting}
        \label{fig:fashion_llc}
    \end{subfigure}

    \caption{LLC and validation loss after splitting, using fashion-MNIST dataset. $\text{SGD lr} = 10^{-2},\text{NGD lr} = 10^{-2}$}
    \label{fig:splitting_fashion}
\end{figure}

\begin{figure}[h]
    \centering
    \begin{subfigure}[t]{0.45\textwidth}
        \includegraphics[width=1.0\textwidth]{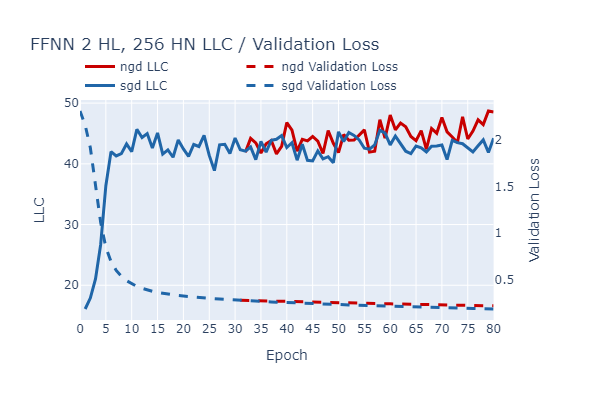}
        \caption{$\text{SGD lr} = 10^{-4},\text{NGD lr} = 10^{-4}$}
        \label{fig:split1e-4}
    \end{subfigure}

    \begin{subfigure}[t]{0.45\textwidth}
        \includegraphics[width=1.0\textwidth]{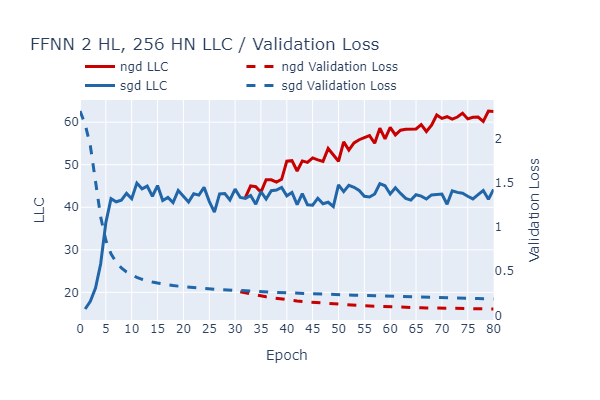}
        \caption{$\text{SGD lr} = 10^{-4},\text{NGD lr} = 10^{-3}$}
        \label{fig:split1e-3}
    \end{subfigure}

    \begin{subfigure}[t]{0.45\textwidth}
        \includegraphics[width=1.0\textwidth]{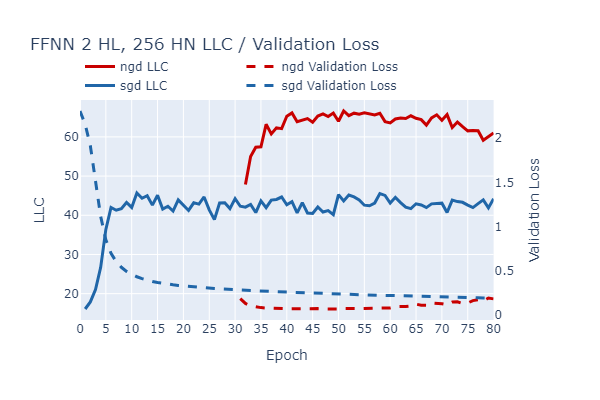}
        \caption{$\text{SGD lr} = 10^{-4},\text{NGD lr} = 10^{-2}$}
        \label{fig:split1e-2}
    \end{subfigure}

    \caption{LLC and validation loss after splitting, using MNIST dataset.}
    \label{fig:splitting_vary_NGD}
\end{figure}

After the split, the update step size of NGD tends to fluctuate while that of SGD is steady, and we often notice an increase in $\hat{\lambda}_{NGD}$  while $\hat{\lambda}_{SGD}$ either remains steady (Figure \ref{fig:splitting_vary_NGD}) or increases at a slower rate (Figure \ref{fig:fashion_llc}). Even if $\hat{\lambda}_{NGD}$ does not increase significantly after the split, increasing the learning rate for NGD makes $\hat{\lambda}_{NGD}$ increase faster and plateau at a higher value as in Figure \ref{fig:splitting_vary_NGD}.

\begin{figure}[h]
    \centering
    \begin{subfigure}[t]{0.45\textwidth}
        \includegraphics[width=1.0\textwidth]{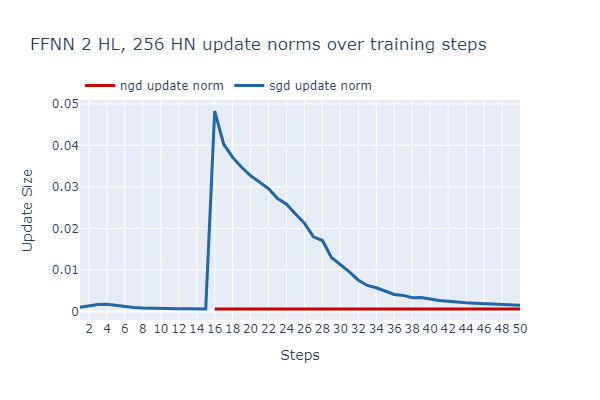}
        \caption{Update step size rises sharply after splitting due to a much higher learning rate. It steadily decreases after.}
    \end{subfigure}

    \begin{subfigure}[t]{0.45\textwidth}
        \includegraphics[width=1.0\textwidth]{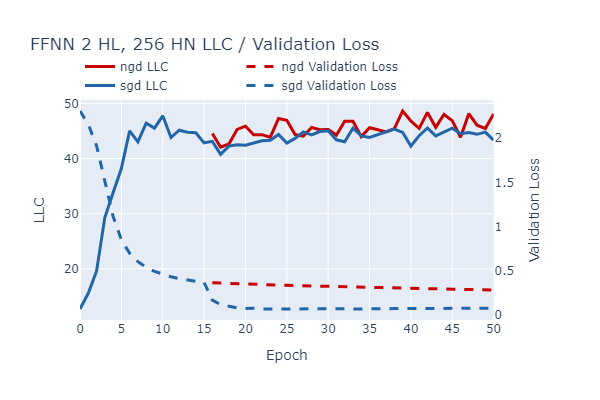}
        \caption{LLC of SGD does not increase even with a sharp rise in step size}
        \label{fig:splitting_sgd_LLC}
    \end{subfigure}

    \caption{Before splitting, SGD lr $= 10^{-4}.$ After splitting, SGD lr is raised to $= 10^{-2}$, while NGD lr $= 10^{-4}$ }
    \label{fig:splitting_SGD}
\end{figure}

We verify whether the increase in $\hat{\lambda}_{NGD}$ is due to the \textit{nature of NGD}, rather than the direct increase in learning rate causing it to break out of a local optimum. We only increase the SGD learning rate after switching and notice that $\hat{\lambda}_{SGD}$ does not increase (Figure \ref{fig:splitting_SGD}). However, the larger step size causes the training and validation loss for SGD to decrease sharply.

This further suggests the rise in $\hat{\lambda}_{NGD}$ is indeed due to the nature of NGD. One explanation is that SGD, after training close to convergence, arrives close to a highly degenerate minimum. After switching to NGD, NGD tends to escape these regions as observed by the higher LLC reflecting less singular points. For the case where both models behave similarly, we believe the NGD learning rate is too low but increasing it will allow $\hat{\lambda}_{NGD}$ to rise much faster and plateau at a higher LLC (Figure \ref{fig:split1e-2}).

\section{Conclusion}

In this paper, we provided evidence that solutions obtained by NGD are less degenerate and more complex, utilizing a greater number of dimensions as indicated by both the LLC and the Hessian trace $\text{Tr}(\mathbf{H})$ . We conducted various experiments to vary the "strength" of NGD and find that this higher complexity is not because $\mathbf{F^{-1}}$ makes the step size larger near degenerate minimums, but rather because $\mathbf{F^{-1}}$ guides the trajectory in the loss landscape away from these degenerate minima.



\normalsize
\bibliography{references}

\appendix
\section*{Appendix}

\section{WBIC}

While the WBIC is usually used for model selection in guiding how the posterior distribution concentrates \citep{Watanabe2013BIC}, it has a similar form to the Bayes generalization error. We provide evidence that the WBIC is correlated with generalization ability.

\subsection{Free Energy}

In Bayesian learning, we wish to find the posterior distribution $ p(w | D_n) $  using Bayes' theorem:

\begin{equation} \label{posterior}
p(w \vert D_n) = \frac{p(D_n \vert w)p(w)}{p(D_n)}
\end{equation}

\begin{itemize}
\item $p(D_n|w)$ : the likelihood of $w$
\item $p(D_n)$ : the probability of observing the data $D_n$, also known as the \textit{marginal likelihood} of $D_n$
\end{itemize}

Since $(X_i, Y_i)$ are i.i.d, we have: 
\begin{equation}
p(D_n \vert w) = \prod_{i=1}^n p(y_i, x_i \vert w) = e^{-nL_n(w)}
\end{equation}

The posterior distribution $p(w|D_n)$ can now be re-written as:

\begin{equation}
p(w|D_n) = \frac{e^{-nL_n(w)}p(w)}{Z_n}
\end{equation}

where $Z_n$ is the partition function (or marginal likelihood)

\begin{equation}
Z_n = p(D_n) = \int_W p(w) \, p(D_n | w) \, dw = \int_W p(w)e^{-nL_n(w)}dw
\end{equation}

The free energy $F_n$ is:

\begin{equation}
F_n = -\log(Z_n)
\end{equation}

\subsection{WBIC and the Bayes Generalization Loss $G_n$}

In the limit as $n \to \infty$, the asymptotic forms of the free energy $F_n$ and the WBIC are the same \citep{Watanabe2013BIC}:

\begin{equation}
WBIC \approx F_n \approx n L_n(w_0) + \lambda \log(n)
\end{equation}

In \citep{lau2023quantifying}, the estimator for the WBIC is used as an estimator for the local free energy, \( F_n \). In SLT, in the limit as \( n \to \infty \), the posterior density will concentrate around models with the lowest $F_n$ hence lowest WBIC. The Bayes generalization loss $G_n$ then tells us how it will perform on new data. $F_n$ and $G_n$ have similar forms: $G_n$ is in fact the "derivative" of $F_n$ and measures the average increase in free energy given new data:

\begin{equation}
G_n = \mathbb{E}_{X_{n+1}}[F_{n+1}] - F_n
\end{equation}

Since WBIC and $G_n$ (Equation \ref{eq:G_n}) have similar forms and both display an accuracy-complexity trade-off, we may expect overfitted models to increase in WBIC and $G_n$.

\subsection{WBIC in overfitted models}

\begin{figure}[h]
    \centering
    \includegraphics[width=0.45\textwidth]{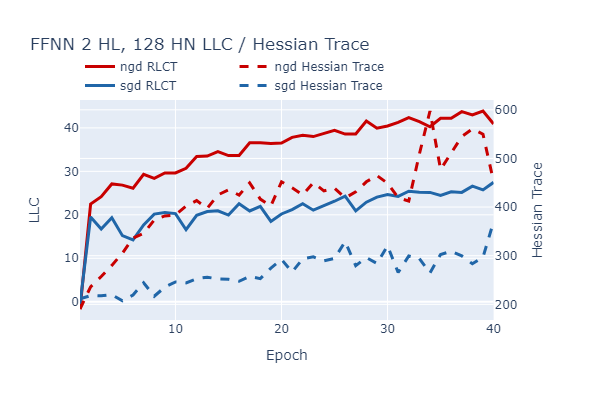}
    \caption{The trace of the Hessian and the LLC increase during overfitting on Fashion MNIST for both SGD and NGD. Hyperparameters used: \( \alpha = 10^{-1}\), \( \text{lr} = 10^{-2}\), \( \epsilon = 10^{-10}\), \( \text{batch size} = 128\).}
    \label{fig:llc_hess}
\end{figure}

\begin{figure}[h]
    \centering
    \includegraphics[width=0.45\textwidth]{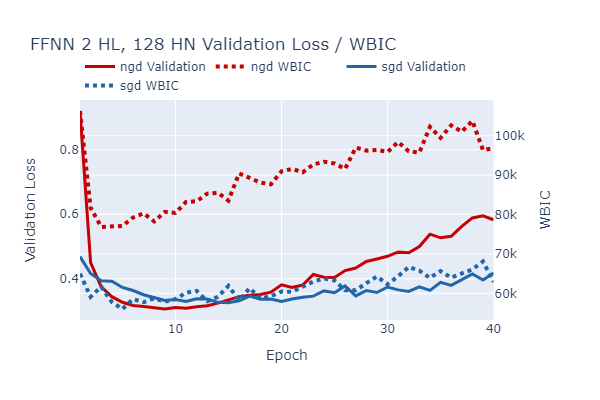}
    \caption{The WBIC increases with validation loss on an overfitted model trained using NGD on Fashion MNIST. Hyperparameters used: \( \alpha = 10^{-1}\), \( \text{lr} = 10^{-2}\), \( \epsilon = 10^{-10}\), \( \text{batch size} = 128\).}
    \label{fig:val_wbic}
\end{figure}

In an overfitted model, training is done for too many epochs, and the model learns noise in the data - this causes poor generalisation on unseen data. We investigated how $\lambda(w^*)$, the WBIC, and \( \text{Tr}(\mathbf{H}) \) evolve, and found that all three quantities increase steadily when the model is overfitting (as validation loss increases). We chose WBIC instead of $G_n$ directly, as only WBIC was implemented in the \texttt{devinterp} library at the point of writing. Figure \ref{fig:llc_hess} shows that the LLC is steadily increasing for both SGD and NGD trained models, and that \( \text{Tr}(\mathbf{H}) \) is also steadily increasing. Figure \ref{fig:val_wbic} shows that the WBIC increases steadily approximately when the model begins to overfit. Intuitively, we should expect $\hat{\lambda}$ and \( \text{Tr}(\mathbf{H}) \) to increase during overfitting, because the model uses more dimensions to capture noise in the data. The WBIC should also increase because it captures the accuracy-complexity tradeoff - during overfitting model complexity increases without increasing its training loss.


Since the WBIC was computed \textit{only using training data}, this hints at the WBIC being a \textit{test-set free} measure of generalization ability.

\section{Computation of the Hessian $\mathbf{H}$}

\subsection{Hessian-vector product $\mathbf{Hv}$}
Computation of the Hessian for a high-parameter model is prohibitive due to its runtime and space (memory) complexity, which are both $O(n^2)$. We therefore compute the trace of the Hessian of the model loss with respect to its parameters $w$, \textit{without explicitly ever computing $\mathbf{H}$}, using the method presented in \citep{yao2020pyhessian}, where they instead compute the Hessian-vector product $\mathbf{H}v$.
\begin{equation}
H = \frac{\partial^2 L}{\partial w^2} = \frac{\partial g_{w}}{\partial w}
\end{equation}
where \( g_w \) is the gradient of the loss \( \nabla L \). We have:

\begin{equation}
\frac{\partial}{\partial w} (g_{w}^T v) = \frac{\partial g_{w}^T}{\partial w} v + g_{w}^T \frac{\partial v}{\partial w}
\end{equation}

Now, considering that $v$ is a random vector and invariant with respect to $w$, we have:

\begin{equation}
\frac{\partial v}{\partial w} = 0
\end{equation}

Note the following:

\begin{equation}
\frac{\partial g_{w}^T}{\partial w} v = \mathbf{H}v
\end{equation}

Therefore, by removing the $\frac{\partial v}{\partial w}$ term:

\begin{equation}
\mathbf{H}v = \frac{\partial}{\partial w} (g_{w}^T v)
\end{equation}

The term $g_{w}^T v$ is a scalar, and so its partial derivative is in $\mathbb{R}^m$ space. In other words, for some vector $v$, we can compute the $\mathbf{Hv}$ in $O(n)$ time; the same as backpropagation.

\subsection{Hessian trace $\text{Tr}(\mathbf{H})$}

Let $v$ be a random vector sampled from a Gaussian distribution with mean $0$ and covariance matrix that is the identity matrix $I$, such that $v \sim \mathcal{N}(\mathbf{0}, \mathbf{I})$. Then, the following, presented in \citep{yao2020pyhessian}, holds:

\begin{equation}
\begin{aligned}
\text{Tr}(\mathbf{H}) &= \text{Tr}(\mathbf{H}I) \\
&= \text{Tr}\left(\mathbf{H} \mathbb{E}(v v^T)\right) \\
&= \mathbb{E}\left( \text{Tr}(\mathbf{H} v v^T)\right) \\
&= \mathbb{E}\left( v^T \mathbf{H} v\right)
\end{aligned}
\end{equation}

Therefore, to compute the trace of $\mathbf{H}$, we just need to sample $N$ random vectors $v$ from the distribution described above, compute $\mathbf{Hv}$ for each of these vectors, and then multiply the result by $v^T$. We found empirically that a value of \( N = 10^{4} \) gave stable, reproducible values of \( \text{Tr}{(\mathbf{H})}\).

\end{document}